\documentclass[runningheads]{llncs}
\usepackage{graphicx,multirow,subcaption}
\usepackage{slashbox}
\usepackage{tabularx, soul, color,amsmath, hyperref}
\usepackage[misc]{ifsym} 

\begin{document}
\title{Deep Placental Vessel Segmentation for Fetoscopic Mosaicking}
%
%
\author{Sophia Bano\inst{1}($^{\textrm{\Letter}}$) \and
Francisco Vasconcelos\inst{1} \and
Luke M. Shepherd\inst{1} \and
Emmanuel Vander Poorten\inst{2} \and
Tom Vercauteren\inst{3} \and
Sebastien Ourselin\inst{3} \and
Anna L. David\inst{4} \and
Jan Deprest\inst{5} \and
Danail Stoyanov\inst{1}}
\authorrunning{S. Bano et al.}
%

\institute{Wellcome/EPSRC Centre for Interventional and Surgical Sciences(WEISS) and Department of Computer Science, University College London, London, UK
\email{sophia.bano@ucl.ac.uk} \and
Department of Mechanical Engineering, KU Leuven University, Leuven, Belgium  \and
School of Biomedical Engineering and Imaging Sciences, King\textsc{\char13}s College London, London, UK  \and
Fetal Medicine Unit, University College London Hospital, London, UK \and
Department of Development and Regeneration, University Hospital Leuven, Leuven, Belgium}
\maketitle              
\begin{abstract}
During fetoscopic laser photocoagulation, a treatment for twin-to-twin transfusion syndrome (TTTS), the clinician first identifies abnormal placental vascular connections and laser ablates them to regulate blood flow in both fetuses. The procedure is challenging due to the mobility of the environment, poor visibility in amniotic fluid, occasional bleeding, and limitations in the fetoscopic field-of-view and image quality. Ideally, anastomotic placental vessels would be automatically identified, segmented and registered to create expanded vessel maps to guide laser ablation, however, such methods have yet to be clinically adopted. We propose a solution utilising the U-Net architecture for performing placental vessel segmentation in fetoscopic videos. The obtained vessel probability maps provide sufficient cues for mosaicking alignment by registering consecutive vessel maps using the direct intensity-based technique. Experiments on 6 different in vivo fetoscopic videos demonstrate that the vessel intensity-based registration outperformed image intensity-based registration approaches showing better robustness in qualitative and quantitative comparison. We additionally reduce drift accumulation to negligible even for sequences with up to 400 frames and we incorporate a scheme for quantifying drift error in the absence of the ground-truth. Our paper provides a benchmark for fetoscopy placental vessel segmentation and registration by contributing the first in vivo vessel segmentation and fetoscopic videos dataset. 

\end{abstract}
\section{Introduction}
Twin-to-twin transfusion syndrome (TTTS) is a rare condition during pregnancy that affects the placenta shared by genetically identical twins~\cite{baschat2011twin}. It is caused by abnormal placental vascular anastomoses on the chorionic plate of the placenta between the twin fetuses that disproportionately allow transfusion of blood from one twin to the other. Fetoscopic laser photocoagulation is a minimally invasive procedure that uses a fetoscopic camera and a laser ablation tool. After insertion into the amniotic cavity, the surgeon uses the scope to identify the inter-twin anastomoses and then photocoagulates them to treat the TTTS. Limited field-of-view (FoV), poor visibility~\cite{LEWI201319}, unusual placenta position~\cite{deprest1998alternative} and limited maneuverability of the fetoscope may hinder the photocoagulation resulting in increased procedural time and incomplete ablation of anastomoses leading to persistent TTTS. 
Automatic segmentation of placental vessels and mosaicking for FoV expansion and creation of a placental vessel map registration may provide computer-assisted interventions (CAI) support for TTTS treatment to support the identification of abnormal vessels and their ablation status. 

CAI techniques in fetoscopy have concentrated efforts on visual mosaicking to create RGB maps of the placental vasculature for surgical planning and navigation~\cite{bano2019deep,daga2016real,peter2018retrieval,reeff2006mosaicing,tella2019pruning}. Several approaches have been proposed for generating mosaics based on: (1) detection and matching of visual point features~\cite{daga2016real,reeff2006mosaicing}; (2) fusion of visual tracking with electromagnetic pose sensing to cater for drifting error in ex vivo experiments~\cite{tella2019pruning}; (3) direct pixel-wise alignment of gradient orientations for a single in vivo fetoscopic video~\cite{peter2018retrieval}; (4) deep learning-based homography estimation for fetoscopic videos captured from various sources~\cite{bano2019deep}; and (5) detection of stable regions using R-CNN and using these regions as features for placental image registration in an underwater phantom setting~\cite{gaisser2018stable}. While promising results have been achieved for short video sequences \cite{bano2019deep}, long-term mapping remains a significant challenge \cite{peter2018retrieval} due to a variety of factors that include occlusion by the fetus, non-planar views, floating amniotic fluid particles and poor video quality and resolution. Some of these challenges can be addressed by identifying occlusion-free views in fetoscopic videos~\cite{bano2020fetnet}. Immersion in the amniotic fluid also causes distortions due to light refraction \cite{chadebecq2017refractive,chadebecq2019refractive} that are often hard to model. Moreover, ground-truth homographies are not available for in vivo fetoscopic videos making qualitative (visual) evaluation the widely used standard for judging the quality of generated mosaics. Quantitative evaluation of registration errors in fetoscopic mosaicking has been limited to ex vivo, phantom, or synthetic experiments.

Fetoscopic mosaicking aims to densely reconstruct the surface of the placenta but limited effort has been directed at identification and localisation of vessels in the fetoscopic video. Placental vasculature can be pre-operatively imaged with MRI for surgical planning \cite{aughwane2019placental}, but there is currently no integration with fetoscopic imaging that enables intra-operative CAI navigation. Moreover, there are no publicly available annotated datasets to perform extensive supervised training. Methods based on a multi-scale vessel enhancement filter~\cite{frangi1998multiscale} have been developed for segmenting vasculature structures from ex vivo high-resolution photographs of the entire placental surface \cite{almoussa2011automated,chang2013vessel}. However, such methods fail on in vivo fetoscopy~\cite{sadda2019deep}, where captured videos have significantly poorer visibility conditions, lower resolution, and a narrower FoV.

Identifying vessels and creating an expanded vessel map can support laser ablation but remains an open problem to date. In this paper, we propose a framework for generating placental vasculature maps that expand the FoV of the fetoscopic camera by performing vessel segmentation in fetoscopic images, registration of consecutive frames, and blending of vessel prediction probability maps from multiple views in a unified mosaic image. We use the U-Net architecture for vessel segmentation since it is robust even with limited training data~\cite{ronneberger2015u}. Comparison with the available alternative~\cite{sadda2019deep} confirmed the superior performance of U-Net for placental vessel segmentation. Alignment of vessel segmentations from consecutive frames is performed via direct registration of the probability maps provided by the U-Net. Finally, multiple overlapping probability maps are blended in a single reference frame. Additionally, we propose the use of quantitative metrics to evaluate the drifting error in sequential mosaicking without relying on the ground-truth (GT). Such temporal evaluation is crucial when GT is not available in surgical video. Our contributions can be summarised as follows:
\begin{itemize}
    \item A placental vessel segmentation deep learning network trained on 483 manually annotated in vivo fetoscopic images that significantly outperforms the available alternative~\cite{sadda2019deep}, showing accurate results on 6 in vivo sequences from different patients, with significant changes in appearance.
    \item Validation of fetoscopic image registration driven exclusively from vessel segmentation maps. We show that, when vessels are visible, this approach is more reliable than performing direct image registration. Many of the visibility challenges in lighting conditions and the presence of moving occlusions are filtered out and vessels are found to have unique recognisable shapes.
    \item A quantitative evaluation of drift registration error for in vivo fetoscopic sequences, by analysing the similarity between overlapping warped images and predicted segmentation maps. This measures the registration consistency after sequential registration of multiple frames.  
    \item Contribute the first placental vessel segmentation dataset (483 images from 6 subjects) and 6 in vivo video clips, useful for benchmarking results in this domain. Completely anonymised videos of TTTS fetoscopic procedure were obtained from the University College London Hospital. This dataset is made publicly available for research purpose here: \url{https://www.ucl.ac.uk/interventional-surgical-sciences/fetoscopy-placenta-data}.

\end{itemize}

\begin{figure}[t]
\centering
\includegraphics[width=1.0\textwidth]{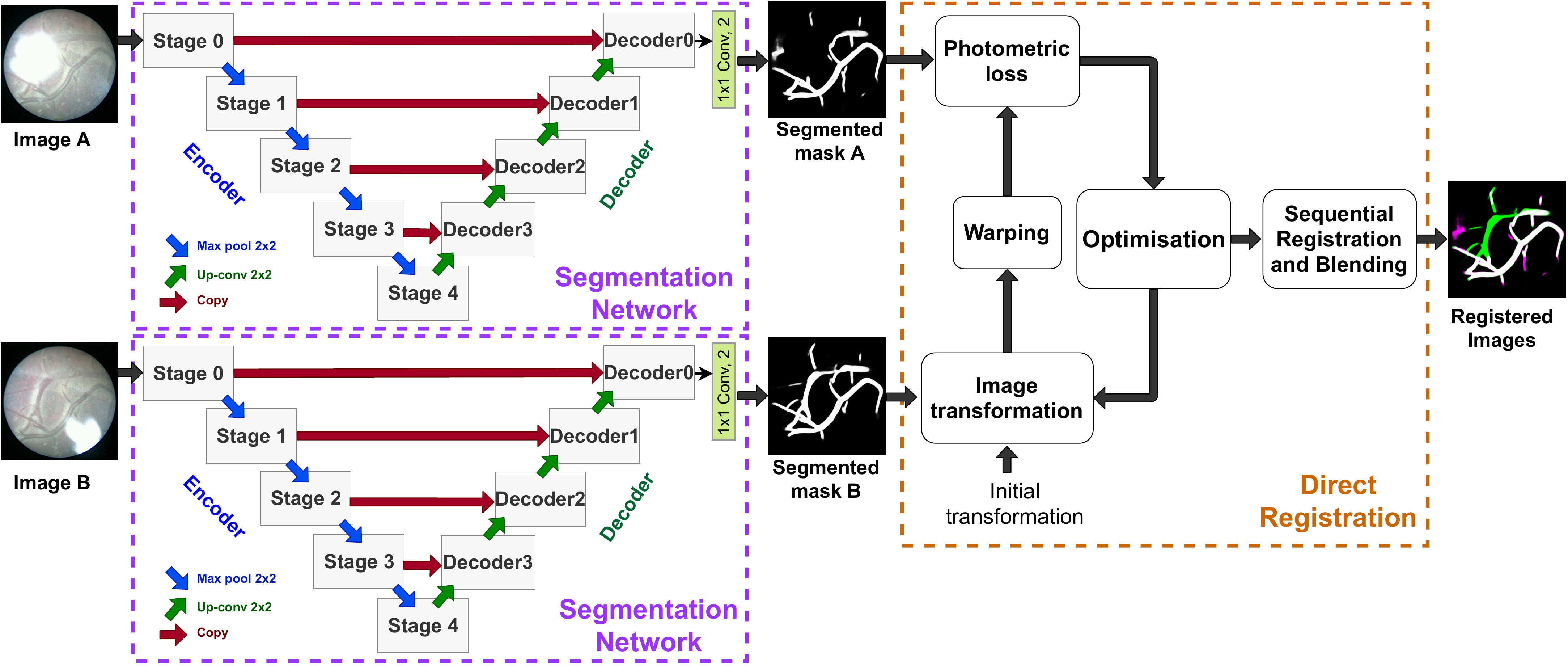}
\caption{An overview of the proposed framework which is composed of the segmentation block followed by the direct registration block for mosaic generation.} 
\label{fig:block_diagram}
\end{figure}

\section{Vessel Segmentation-based Mosaicking}
Our proposed framework consists of a segmentation block followed by the registration block as shown in Fig.~\ref{fig:block_diagram}. A U-Net architecture is used for obtaining the prediction maps for the vessels (Sec.~\ref{sec:vessel_seg}). Vessel probability maps from two consecutive frames are then aligned through affine registration (Sec.~\ref{sec:registration}). These transformations are accumulated in sequence with respect to the first frame to generate an expanded view of the placental vessels. 

\subsection{U-Net for Placental Vessel Segmentation}
\label{sec:vessel_seg}
U-Net~\cite{ronneberger2015u} provides the standard architecture for semantic segmentation and continues to be the base of many of the state-of-the-art segmentation models~\cite{badrinarayanan2017segnet,litjens2017survey}. 
U-Net is a fully convolutional network which is preferred for medical image segmentation since it results in accurate segmentation even when the training dataset is relatively small. Placental vessel segmentation is considered as a pixel-wise binary classification problem. Unlike~\cite{ronneberger2015u}, which used the binary cross-entropy loss ($L_{bce}$), we use the sum of binary cross-entropy loss and intersection over union (Jaccard) loss during training given by: 
\begin{equation}
\small
\begin{aligned}
    L(p,\hat{p}) &= L_{bce}(p,\hat{p})+L_{iou}(p,\hat{p}),\\
    &= -\frac{\sum{\left [p(\log \hat{p})+(1-p)\log (1-\hat{p})\right]}}{N}+\left [1-\frac{\sum (p\,.\, \hat{p})+\delta }{\sum(p + \hat{p})-\sum (p\,.\, \hat{p})+\delta}\right ],
\end{aligned}  
\normalsize
\label{eq:1}
\end{equation}
where $p$ is the flattened ground-truth label tensor, $\hat{p}$ is the flattened predicted label tensor, $N$ is the total number of pixels in the image and $\delta=10^{-5}$ is arbitrary selected to avoid division by zero. We empirically found that the combined loss (eq.~\ref{eq:1}) results in improved accuracy compared to when $L_{bce}$ alone is used. Detailed description of the U-Net can be found in \cite{ronneberger2015u}. The vessel probability maps from the U-Net are then used for the vessel registration.  

\subsection{Vessel Map Registration and Mosaicking}
\label{sec:registration}
Segmentations from consecutive frames are aligned via registration of the probability maps provided by the segmentation network. We perform registration on the probability maps, rather than on binary segmentation masks as this provides smoother cost transitions for iterative optimisation frameworks, facilitating convergence. We approximate the registration between consecutive frames with affine transformations. We confirmed the observations in \cite{peter2018retrieval} that projective registration leads to worse results. In our in vivo datasets, camera calibration is not available and lens distortion cannot be compensated, therefore the use of projective transformations can lead to excessive over-fitting to distorted patterns on the image edges. Existing literature~\cite{peter2018retrieval} shows that direct-registration approach is robust in fetoscopy where feature-based methods fail due to lack of texture and resolution~\cite{bano2019deep}. Therefore, we use a standard pyramidal Lucas-Kanade registration framework \cite{bouguet2001pyramidal} that minimises the bidirectional least-squares difference, also referred to as photometric loss, between a fixed image and a warped moving image. A solution is found with the Levenberg-Marquardt iterative algorithm. Several available implementations of Levenberg-Marquardt offer the option of computing the optimisation step direction through finite differences (e. g. MATLAB's lsqnonlin implementation), however, we have observed that explicitly computing it using the Jacobian of the cost function is necessary for convergence. Given the fetoscopic images have a circular FoV, the registration is performed with a visibility mask and a robust error metric is applied to deal with complex overlap cases \cite{brunet2010direct}. 

After multiple sequential registrations are performed, they can be blended into a single image, which not only expands the FoV of the vessel map but also filters single-frame segmentation errors whenever multiple results overlap. This is achieved by taking the average probability over all available overlapping values for each pixel, leading to the results presented in Sec.~\ref{sec:results}.

\section{Experimental Setup and Evaluation Protocol}
\begin{figure}[t]
	\centering
	\includegraphics[width=1.0\textwidth]{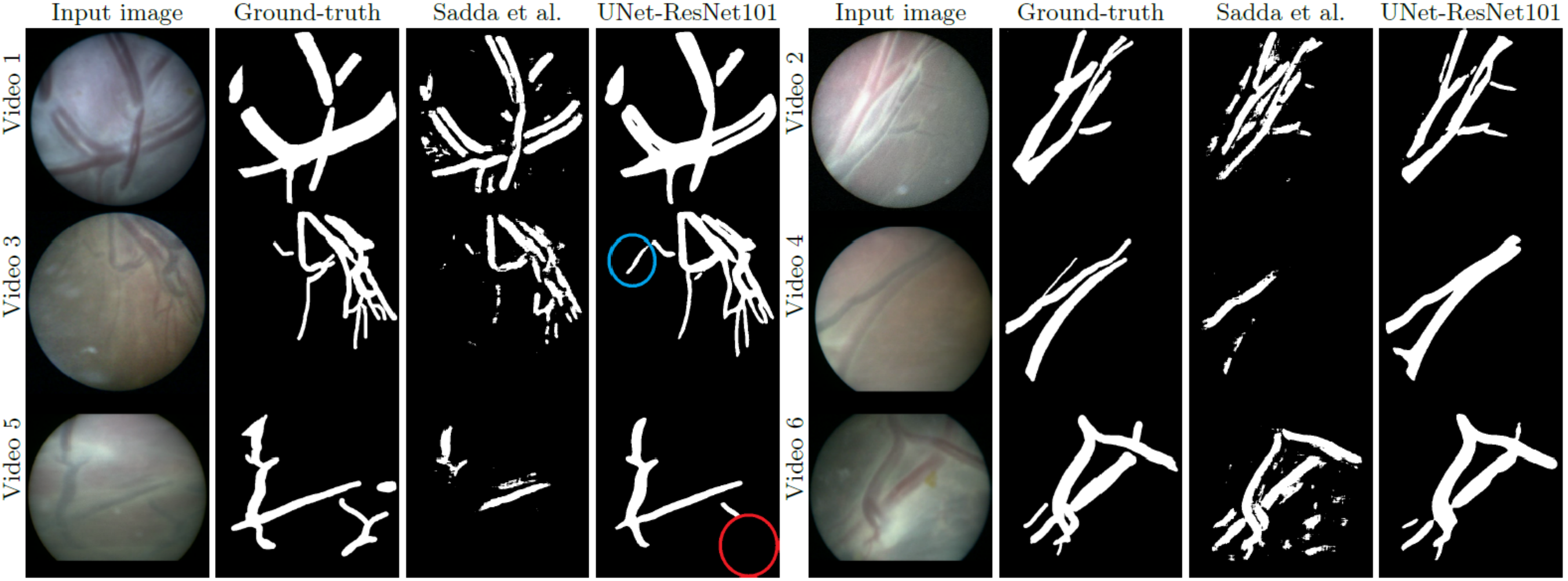}
	\caption{Qualitative comparison of U-Net (ResNet101) with the baseline (Sadda et al.~\cite{sadda2019deep}) network for the placental vessel segmentation.} 
	\label{fig:seg_qualitative}
\end{figure}

We annotated 483 sampled frames from 6 different in vivo TTTS laser ablation videos. The image appearance and quality varied in each video due to the variation in intra-operative environment among different cases, artefacts and lighting conditions, resulting in increased variability in the data (sample images are shown in Fig.~\ref{fig:seg_qualitative}). We first selected the non-occluded (no fetus or tool presence) frames through a separate frame-level fetoscopic event identification approach~\cite{bano2020fetnet} since vessels are mostly visible in such frames. The videos were then down-sampled from 25 to 1 \textit{fps} to avoid including very similar frames in the annotated samples and each frame was resized to $448\times448$ pixel resolution. The number of clear view samples varied in each in vivo video, hence the number of annotated images varied in each fold (Table~\ref{tab:seg_quantitative}). We use the pixel annotation tool\footnote{Pixel annotation tool: \url{https://github.com/abreheret/PixelAnnotationTool}} for annotating and creating a binary mask of vessels in each frame. A fetal medicine specialist further verified our annotations to confirm the correctness of our labels. For direct image registration, we use continuous unannotated video clips from the 6 in vivo videos. 

We perform 6-fold cross-validation to compare and verify the robustness of the segmentation algorithms. Mean Intersection over Union (IoU) and Dice (F1) scores are used to evaluate the segmentation performance (reported in Table~\ref{tab:seg_quantitative}). We experiment with the vanilla U-Net~\cite{ronneberger2015u} and with VGG16~\cite{simonyan2014very}, ResNet50~\cite{he2016deep} and ResNet101~\cite{he2016deep} backbones (with pre-trained Imagenet weights) to search for the best performing architecture. In each training iteration, a sub-image of size $224\times224$ is cropped at random after augmenting the image with rotation, horizontal or vertical flip, and illumination intensity change randomly. This helps in increasing the data and variation during training. A learning rate of $3e^{-4}$ with Adam optimiser and our combined loss (eq.~\ref{eq:1}) is used. For each fold, training is performed for 1000 epochs with early stopping and the best performing weights on the training dataset are captured and used to validate the performance on the left-out (hold) set of frames. 

\begin{table}[t!]
	\centering
	\caption{Six fold cross-validation and comparison of the baseline~\cite{sadda2019deep} with U-Net architecture (having different backbones) for placental vessel segmentation.}
	\label{tab:seg_quantitative}
	\resizebox{1.0\textwidth}{!}{
	\footnotesize
	\begin{tabular}{|l|l||c|c|c|c|c|c||c|}
		\hline 
		\multicolumn{2}{|l||}{} &\textbf{Fold 1} &\textbf{Fold 2} &\textbf{Fold 3} &\textbf{Fold 4} &\textbf{Fold 5}
		&\textbf{Fold 6} &\textbf{Overall}    \\ \hline 
		\multicolumn{2}{|l||}{\textbf{No. of validation images}} &121 &101 &39 &88 &37 &97 &483  \\ \hline
		\textbf{Sadda et al.\cite{sadda2019deep}} &Dice &$0.61\pm0.17$ &$0.57\pm0.17$ &$0.54\pm0.17$ &$0.59\pm0.17$ &$0.56\pm0.16$ &$0.50\pm0.16$ &$0.57\pm0.17$ \\ \cline{2-9}
		(Baseline) &IoU &$0.46\pm0.16$ &$0.42\pm0.15$ &$0.39\pm0.16$ &$0.43\pm0.16$ &$0.40\pm0.15$ &$0.35 \pm0.16$ &$0.41\pm0.16$\\ \hline 
		\textbf{U-Net (Vanilla)\cite{ronneberger2015u}} &Dice &$0.82\pm0.12$ &$0.73\pm0.17$ &$0.82\pm0.09$ &$0.70\pm0.20$ &$0.67\pm0.19$ &$0.72\pm0.13$ &$0.75\pm0.15$ \\ \cline{2-9}
		&IoU &$0.71\pm0.13$ &$0.60\pm0.18$ &$0.70\pm0.19$ &$0.57\pm0.22$ &$0.53\pm0.19$ &$0.58\pm0.15$ &$0.62\pm0.17$\\ \hline 
		\textbf{U-Net (VGG16)} &Dice &$0.82\pm0.12$ &$0.69\pm0.14$ &$\textbf{0.84}\pm\textbf{0.08}$ &$0.73\pm0.19$ &$0.70\pm0.18$ &$0.74\pm0.14$ &$0.75\pm0.14$ \\ \cline{2-9}
		&IoU &$0.71\pm0.14$ &$0.55\pm0.16$ &$\textbf{0.73}\pm\textbf{0.11}$ &$0.61\pm0.21$ &$0.56\pm0.19$ &$0.60\pm0.16$ &$0.63\pm0.16$\\ \hline 
		\textbf{U-Net (Resnet50)} &Dice &$0.84\pm0.10$ &$0.74\pm0.14$ &$0.83\pm0.09$ &$0.74\pm0.19$ &$\textbf{0.72}\pm\textbf{0.17}$ &$0.72\pm0.16$ &$0.77\pm0.14$ \\ \cline{2-9}
		&IoU &$\textbf{0.74}\pm\textbf{0.12}$ &$0.61\pm0.16$ &$\textbf{0.73}\pm\textbf{0.12}$ &$\textbf{0.62}\pm\textbf{0.21}$ &$\textbf{0.58}\pm\textbf{0.18}$ &$0.58\pm0.17$ &$0.65\pm0.16$\\ \hline
		\textbf{U-Net (Resnet101)} &Dice &$\textbf{0.85}\pm\textbf{0.07}$ &$\textbf{0.77}\pm\textbf{0.16}$ &$0.83\pm0.08$ &$\textbf{0.75}\pm\textbf{0.18}$ &$0.70\pm0.18$ &$\textbf{0.75}\pm\textbf{0.12}$ &$\textbf{0.78}\pm\textbf{0.13}$ \\ \cline{2-9}
		&IoU &$\textbf{0.74}\pm\textbf{0.10}$ &$\textbf{0.64}\pm\textbf{0.17}$ &$0.72\pm0.12$ &$\textbf{0.62}\pm\textbf{0.20}$ &$0.56\pm0.19$ &$\textbf{0.62}\pm\textbf{0.15}$ &$\textbf{0.66}\pm\textbf{0.15}$ \\     \hline
	\end{tabular}
	}
\end{table}

In order to evaluate our segmentation driven registration approach, we compare it against standard intensity-based registration. We use the same registration pipeline (described in Sec.~\ref{sec:registration}) for both approaches, only changing the input data. Quantification of fetoscopic mosaicking performance with in vivo data remains a difficult challenge in the absence of GT registration. We propose to indirectly characterise the drift error accumulated by consecutive registrations by aligning non-consecutive frames with overlapping FoVs and measuring their structural similarity (SSIM) and the IoU of their vessel predicted maps (results shown in Fig.~\ref{fig:quant_comparison}). We use a non-overlapping sliding window of 5 frames and compute the SSIM and IoU between frame 1 and the reprojected non-consecutive frames (from 2 to 5). We highlight that this evaluation is mostly suitable for strictly sequential registration, when non-consecutive frames can provide an unbiased measure of accumulated drift.

\section{Results and Discussion}
\label{sec:results}
We perform comparison of the U-Net (having different backbones) with the existing baseline placental vessel segmentation network~\cite{sadda2019deep}. The qualitative comparison is shown in Fig.~\ref{fig:seg_qualitative} and 6-fold cross-validation comparison is reported in Table~\ref{tab:seg_quantitative}. Sadda~et~al.~\cite{sadda2019deep} implemented a 3-stage U-Net with 8 convolutions in each stage and designed a positive class weighted loss function for optimisation. Compared to the Vanilla U-Net (IoU $=0.62$), the performance of \cite{sadda2019deep} is lower ((IoU $=0.41$). When experimenting with U-Net with different backbones, we found that the results are comparable though U-Net with ResNet101 backbone overall showed an improvement of $3\%$ Dice ($4\%$ IoU) over the Vanilla U-Net. Therefore, we selected U-Net with ResNet101 as the segmentation architecture for obtaining vessel probability maps for completely unlabelled clips from the 6 fetoscopic videos. The robustness of the obtained vessel probability maps is further verified by the generated mosaics. From Table~\ref{tab:seg_quantitative}, we note that the segmentation results on fold-5 are significantly lower for all methods. This is mainly because most of the vessels were thin in this validation set and there were a few erroneous manual annotations which were rightly detected as false negative. From Fig.~\ref{fig:seg_qualitative} (video 3, blue circle), we observe that U-Net (ResNet101) even managed to detect small vessels which were originally missed by the human annotator. In video 5, some vessels around the lower right side of the image were missed due to poor illumination (indicated by red circle).    

\begin{figure}[t]
    \centering
    \subcaptionbox{Structural Similarity Index (SSIM)\label{fig:SSIM}}
    {\includegraphics[width=\textwidth]{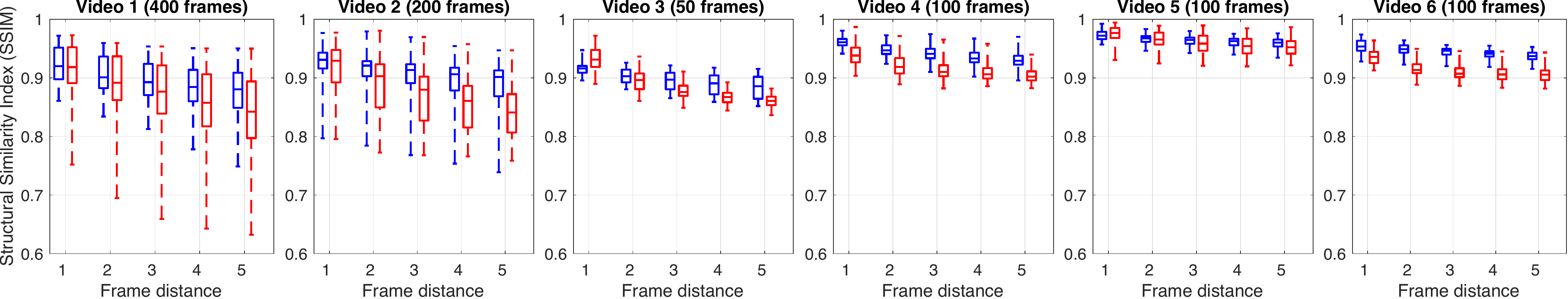}}
    \subcaptionbox{Intersection over Union (IoU)\label{fig:IoU}}
    {\includegraphics[width=\textwidth]{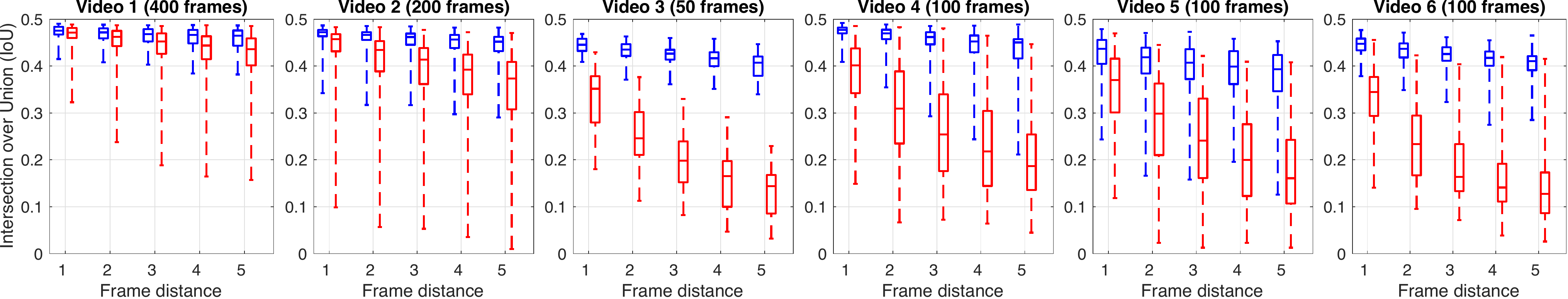}}
    \caption{Registration performance while using vessel prediction maps (blue) and intensity images (red) as input over 5 frame distance.}
    \label{fig:quant_comparison}
\end{figure}

\begin{figure}[t!]
    \centering
    \subcaptionbox{Vessel-based Registration \label{fig:mosaicseg}}
    {\includegraphics[height=37mm]{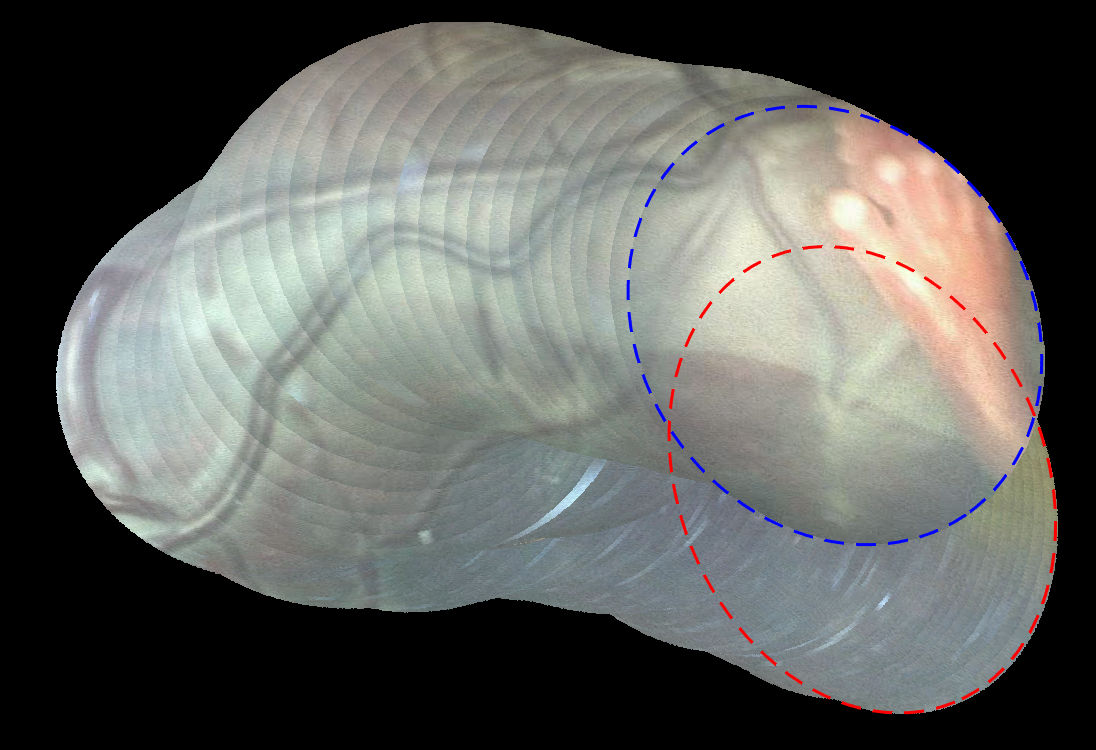}}
    \subcaptionbox{Image-based registration \label{fig:mosaicimg}}
    {\includegraphics[height=37mm]{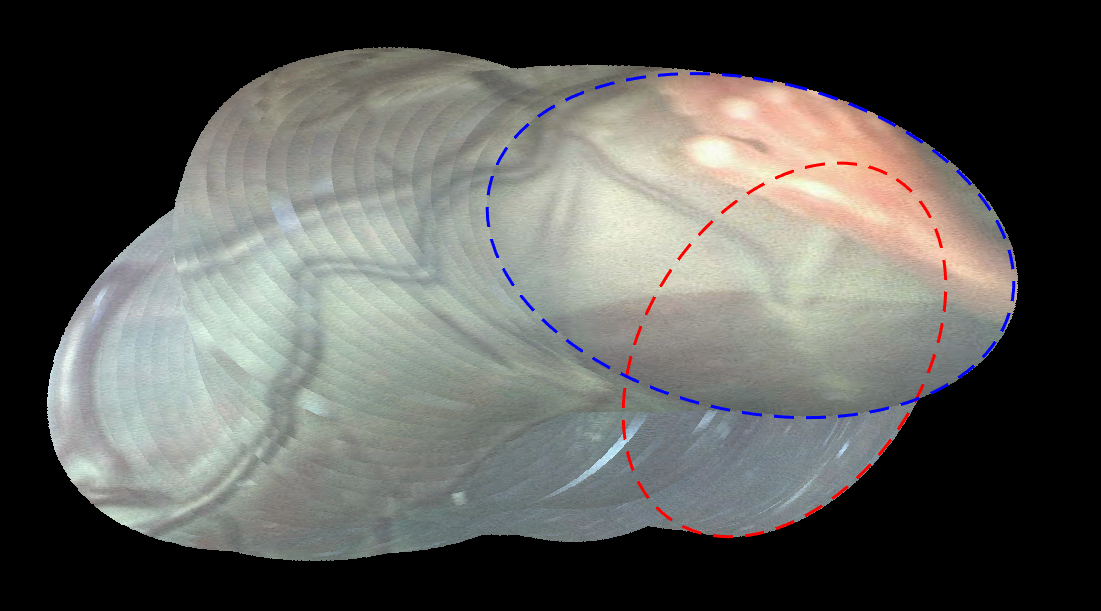}}
    \caption{Image registration on video 1 (400 frames duration). First (blue) and last (red) frames are highlighted. Mosaics are generated relative to the frame closest to the centre (i. e. this frame has the same shape as the original image).}
    \label{fig:vessel_vs_gray}
\end{figure}

\begin{figure}[t]
    \centering
    \includegraphics[width=\textwidth]{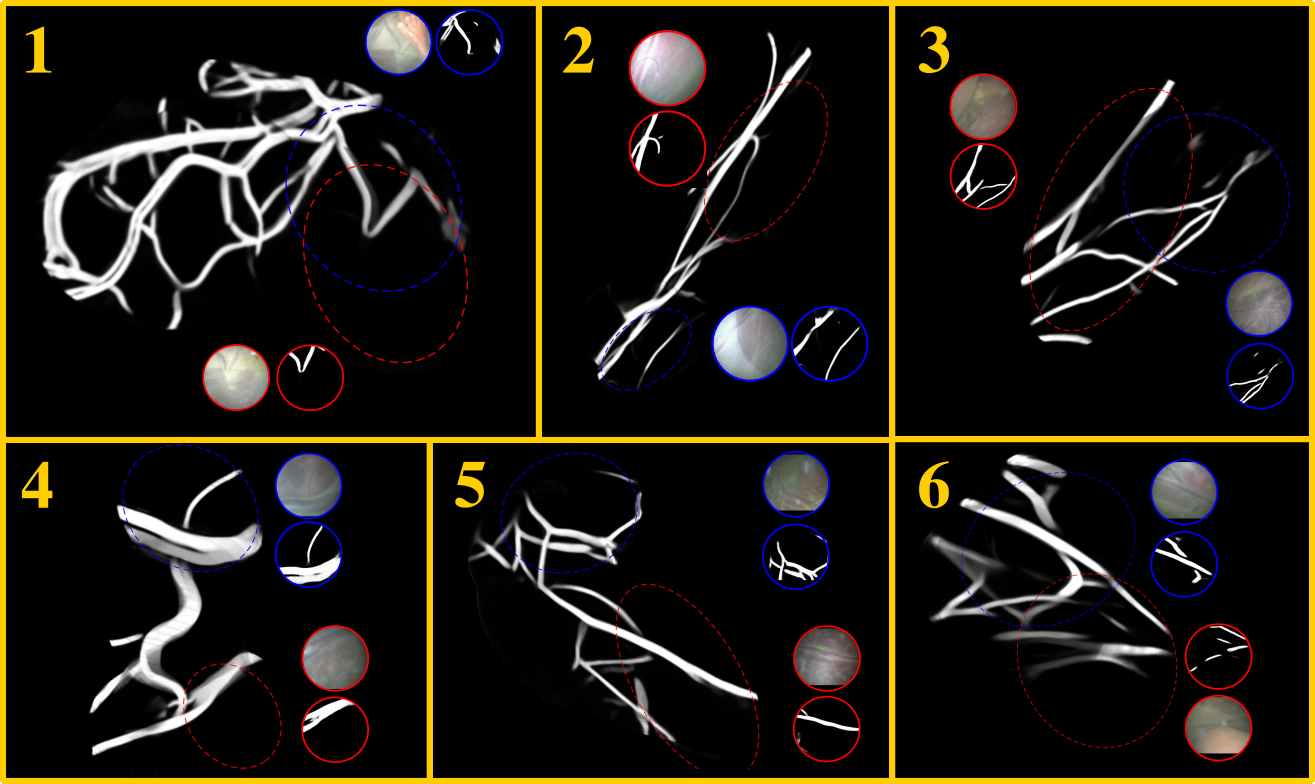}  
    \caption{Visualisation of the vessel maps generated from the segmentation predictions for the 6 in vivo clips. First (blue) and last (red) frames are highlighted. Refer to the \href{https://youtu.be/BU66ctiUUj0}{supplementary video} for the qualitative comparison.}
    \label{fig:vessel_maps}
\end{figure}

The performance comparison between the vessel-based and image-based registration shown in Fig.~\ref{fig:quant_comparison} reports the SSIM and IoU scores for all image/ prediction-map pair overlaps that are up to 5 frames apart. For each of the 6 unseen and unlabelled test video clips, we use the segmentation network trained on the frames from the remaining 5 videos as input for predicting the vessel maps and registration. It is worth noting that our method has usually lower SSIM values than the baseline between consecutive frames (leftmost box of plots in Fig. \ref{fig:SSIM}). This is to be expected since the baseline overfits to occlusions in the scene (amniotic fluid particles, fetus occlusions, etc). However, when analysing the overlap 5 frames apart our method becomes more consistent. Note from Fig.\ref{fig:IoU} (Video 3 to 6) the gradually decreasing IoU over the 5 consecutive frames for the image-based registration compared to the vessel-based registration. This effect is because of the increasing drift in the image-based registration method resulting in heavy misalignment. The qualitative comparison in Fig.~\ref{fig:vessel_vs_gray} further supports our method's better performance. Using vessel prediction maps are more robust since it overcomes challenges such as poor visibility and resolution that accounts for introducing drift in the image-based registration. Figure~\ref{fig:vessel_maps} shows the qualitative results of the vessel-based registration for the 6 leave-one-out unlabelled in vivo video clips (refer to the \href{https://youtu.be/BU66ctiUUj0}{supplementary video} for the sequential qualitative comparison). Note that vessel-based mosaicking not only generated an increased FoV image but also helped in improving vessel segmentation (occluded or missed vessel) results by blending several registered frames.  

\subsubsection{Acknowledgments}This work was supported by the Wellcome/EPSRC Centre for Interventional and Surgical Sciences (WEISS) at UCL (203145Z/16/Z), EPSRC (EP/P027938/1, EP/R004080/1,NS/A000027/1), the H2020 FET (GA 863146) and Wellcome [WT101957]. Danail Stoyanov is supported by a Royal Academy of Engineering Chair in Emerging Technologies (CiET1819/2/36) and an EPSRC Early Career Research Fellowship (EP/P012841/1). Tom Vercauteren is supported by a Medtronic$/$Royal Academy of Engineering Research Chair [RCSRF1819/7/34].

\section{Conclusion}
We proposed a placental vessel segmentation driven framework for generating chorionic plate vasculature maps from in vivo fetoscopic videos. Vessel probability maps were created by training a U-Net on 483 manually annotated placental vessel images. Direct vessel-based registration was performed using the vessel probability maps which not only helped in minimising error due to drift but also corrected missing vessels that occurred due to partial occlusion in some frames, alongside providing a vascular map of the chorionic plate of the mono-chorionic placenta. The proposed framework was evaluated through both quantitative and qualitative comparison with the existing methods for validating the segmentation and registration blocks. Six different in vivo video clips were used to validate the generated mosaics. Our proposed framework along with the contributed vessel segmentation and in vivo fetoscopic videos datasets provide a benchmark for future research on this problem.

\bibliographystyle{splncs04}
\bibliography{Bib_FetVessel_Reg}

\end{document}